\def\year{2018}\relax
\documentclass[letterpaper]{article} 
\usepackage{aaai18}  
\usepackage{times}  
\usepackage{helvet}  
\usepackage{courier}  
\usepackage{url}  
\usepackage{graphicx}  
\usepackage[ruled,vlined]{algorithm2e}
\usepackage{amsmath}
\usepackage{amsfonts}
\usepackage{blindtext}
\usepackage{color}
\begin{document}
%
\title{Order-Free RNN with Visual Attention
for Multi-Label Classification}
\author{Shang-Fu Chen$^1$\thanks{ - indicates equal contribution.}, Yi-Chen Chen$^2$\footnotemark[1], Chih-Kuan Yeh$^3$, Yu-Chiang Frank Wang$^{1,2}$\\
$^1$Department of Electrical Engineering, National Taiwan University, Taipei, Taiwan\\
$^2$Graduate Institute of Communication Engineering, National Taiwan University, Taipei, Taiwan\\
$^3$Machine Learning Department, Carnegie Mellon University, Pittsburgh, USA\\
Email: \{b02901030, r06942069, ycwang\}@ntu.edu.tw, cjyeh@cs.cmu.edu
}
\maketitle
\def \D {\textbf{D}} 
\def \R {\mathbb{R}}

\def \X {\textbf{X}} 
\def \Y {\textbf{Y}} 

\def \x {\textbf{x}} 
\def \y {\textbf{y}} 
\def \hx {\hat{\x}} 
\def \hy {\hat{\y}} 
\def \tl {l} 
\def \ty {\tilde{\y}} 

\def \c {c} 
\def \C {\mathcal{C}} 
\def \pC {\C'}

\def \f {\mathbf{M}} 
\def \fcnn {\f_{map}}
\def \fatt {\f_{att}}
\def \flstm {\f_{pred}}

\def \I {\mathbf{I}} 
\def \V {\mathbf{V}} 
\def \z {\mathbf{z}} 
\def \p {\mathbf{p}} 
\def \h {\mathbf{h}} 

\def \v {\mathbf{v}} 
\def \one {\mathbf{1}}

\def \L {\mathbf{L}} 

\section{Abstract}
\label{sec:abstract}
We propose a recurrent neural network (RNN) based model for image multi-label classification. Our model uniquely integrates and learning of visual attention and Long Short Term Memory (LSTM) layers, which jointly learns the labels of interest and their co-occurrences, while the associated image regions are visually attended. Different from existing approaches utilize either model in their network architectures, training of our model does not require pre-defined label orders. Moreover, a robust inference process is introduced so that prediction errors would not propagate and thus affect the performance. Our experiments on NUS-WISE and MS-COCO datasets confirm the design of our network and its effectiveness in solving multi-label classification problems.

\section{Introduction}
\label{sec:intro}

Multi-label classification has been an important and practical research topic, since it needs to assign more than one label to each observed instance. From machine learning, data mining, and computer vision, a variety of applications benefit from the development and success of multi-label classification algorithms~\cite{zhang2014review,boutell2004learning,schapire2000boostexter,godbole2004discriminative,lin2014microsoft,kang2016object,kang2016t,boutell2004learning,shao2016slicing}. A fundamental and challenging issue for multi-label classification is to identify and recover the co-occurrence of multiple labels, so that satisfactory prediction accuracy can be expected.

Recently, development of deep convolutional neural networks (CNN)~\cite{krizhevsky2012imagenet,szegedy2015going,simonyan2014very,he2016deep} have made a remarkable progress in several research fields. Due to its ability of representation learning with prediction guarantees, CNNs contribute to the recent success in image classification tasks and beyond~\cite{deng2009imagenet,fei2007learning,griffin2007caltech}.
Despite its effectiveness, how to extend CNNs for solving multi-label classification problems is still a research direction to explore.

While a number of research works~\cite{zhang2006multilabel,nam2014large,gong2013deep,wei2014cnn,wang2016cnn} start to advance the CNN architecture for multi-label classification, CNN-RNN~\cite{wang2016cnn} embeds image and semantic structures by projecting both features into a joint embedding space. By further utilizing the component of Long Short Term Memory (LSTM)~\cite{hochreiter1997long}, a recurrent neural network (RNN) structure is introduced to memorize long-term label dependency. As a result, CNN-RNN exhibits promising multi-label classification performance with cross-label correlation implicitly preserved.


\begin{figure*}[t!]
	\centering
	\includegraphics[width=0.95\textwidth]{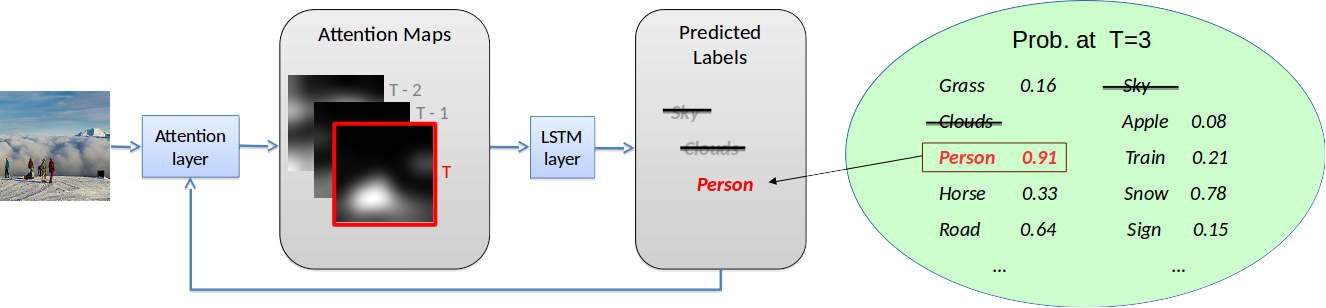}
	\caption{Illustration of our proposed model for multi-label classification. Note that the joint learning of attention and LSTM layers allows us to identify the label dependency without using any predetermined label order, while the corresponding image regions of interest can be attended to accordingly.}
	\label{fig:1}
\end{figure*}

Unfortunately, the above frameworks suffer from the following three different problems. First, due to the use of LSTM, a pre-defined label order is required during training. Take~\cite{wang2016cnn} for example, its label order is determined by the frequencies of labels observed from the training data. In practice, such pre-defined orders of label prediction might not reflect proper label dependency. For example, based on the number of label occurrences, one might obtain the label sequence as \{\emph{sea, sun, fish}\}. However, it is obvious that \emph{fish} is less semantically relevant to \emph{sun} than \emph{sea}. For better learning and prediction of such labels, the order of \{\emph{sea, fish, sun}\} should be considered. On the other hand,~\cite{jin2016annotation} consider four experimental settings with different label orders: alphabetical order, random order, frequency-first order and rare-first order (note that rare-first is exactly the reverse of frequency-first). It is concluded in~\cite{jin2016annotation} that the rare-first order results in the best performance. Later we will conduct thorough experiments for verification, and show that orders automatically learned by our model would be desirable.

The second concern of the above methods is that, labels of objects which are in smaller scales/sizes in images would often be more difficult to be recovered. As a possible solution, attention map~\cite{xu2015show} has been widely considered in image captioning \cite{xu2015show}, image question answering \cite{yang2016stacked}, and segmentation \cite{hong2016learning}. Extracted by different kernels from a certain convolutional layer in CNN, the corresponding feature maps contain rich information of different patterns from the input image. By further attending on such feature maps, the resulting attention map is able to identify important components or objects in an image. By exploiting the label co-occurrence between the associated objects in different scales or patterns, the above problem can be properly alleviated. However, this technique could not be easily applied to RNN-based methods for multi-label problems. As noted above, such methods determine the label order based on the occurrence frequency. For example, the class \emph{person} may appear more often than \emph{horse} in an image collection, and thus the label sequence would be derived as \{\emph{person, horse}\}. Even if the image region of \emph{horse} is typically larger than that of \emph{person}, it might not assist in identifying the rider on its back (i.e., requiring the prediction order as \{\emph{horse, person}\}).

Thirdly, inconsistency between training and testing procedures would often be undesirable for solving multi-label classification tasks. To be more precise, during the training phase, the labels to be produced at each recurrent layer is selected from the ground truth list during the training phase; however, the labels to be predicted during testing are selected from the entire label set. In other words, if a label is incorrectly predicted during a time step during prediction, such an error would propagate during the recurrent process and thus affect the results.

To resolve the above problems, we present a novel deep learning framework of visually attended RNN, which consists of visual attention and LSTM models as shown in Fig.~\ref{fig:1}. In particular, we propose a confidence-ranked LSTM which reflects the label dependency with the introduced visual attention model. Our joint learning framework with the introduced attention model allows us to identify the regions of interest associated with each label. As a result, the order of labels can be automatically learned without any prior knowledge or assumption. As verified later in the experiments, even the objects are presented in small scales in the input image, the corresponding image regions would still be visually attended. More importantly, our network architecture can be applied to both training and testing, and thus the aforementioned inconsistency issue is addressed.


The main contributions of this paper are listed below: 
\begin{itemize}
	\item Without pre-determining the label order for prediction, our method is able to sequentially learn the label dependency using the introduced LSTM model.
	\item The introduced attention model in our architecture allows us to focus on image regions of interests associated with each label, so that improved prediction can be expected even if the objects are in smaller sizes.
	\item By jointly learning attention and LSTM models in a unified network architecture, our model performs favorably against state-of-the-art deep learning approaches on multi-label classification, even if the ground truth label might not be correctly presented during training.
\end{itemize}

\section{Related Work}
\label{sec:rewo}
We first review the development of multi-label classification approaches. Intuitively, the simplest way to deal with multi-label classification problems is to decompose them into multiple binary classification tasks~\cite{tsoumakas2006multi}. Despite its simplicity, such techniques cannot identify the relationship between labels.

To learn the interdependency between labels for multi-label classification, approaches based on classifier chains~\cite{read2011classifier} were proposed, which capture label dependency by conditional product of probabilities. However, in addition to high computation cost when dealing with a larger number of labels, classifier chains have limited ability to capture the high order correlations between labels. On the other hand, probabilistic graphical model based methods  \cite{li2014multi,li2016conditional} learn label dependencies with graphical structure, and latent space methods \cite{yeh2017learning,bhatia2015sparse} choose to project features and labels into a common latent space. Approaches like \cite{yang2016exploit} further utilize additional information like bounding box annotations for learning their models.

\begin{figure*}[t!]
	\centering
	\includegraphics[width=0.95\textwidth]{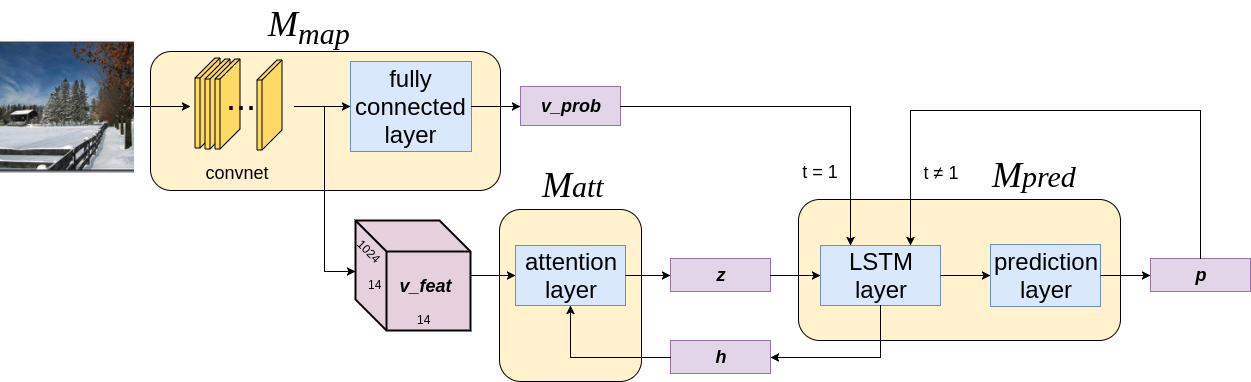}
	\caption{Architecture of our proposed network architecture for multi-label classification. Note that $\fcnn$, $\fatt$, and $\flstm$ indicate the layers for feature mapping, attention, and label prediction, respectively. $v_{feat}$ is a set of feature maps extracted from $\fcnn$, and the vector output $v_{prob}$ represents the preliminary label prediction of $\fcnn$ to initiate LSTM prediction. $z$ and $h$ are the attention context vector and the LSTM hidden state, respectively. Finally, $p$ denotes the vector output indicating the label probability, updated at every time step.}
	\label{fig:2}
\end{figure*}

With the recent progress of neural networks and deep learning, BP-MLL~\cite{zhang2006multilabel} is among the first to utilize neural network architectures to solve multi-label classification. It views each output node as a binary classification task, and relies on the architecture and loss function to exploit the dependency across labels. It was later extended by~\cite{nam2014large} with state-of-the-art learning techniques such as dropout.

Furthermore, state-of-the-art DNN based multi-label algorithms have proposed different loss functions or architectures \cite{gong2013deep,wei2014cnn,hu2016learning}. For example, Gong et al.~\cite{gong2013deep} design a rank-based loss and compensate those with lowest ranks ones, Wei et al.~\cite{wei2014cnn} generate multi-label candidates on several grids and combine results with max-pooling, and Hu et al. propose structured inference NN \cite{hu2016learning}, which uses concept layers modeled with label graphs.

Recurrent neural networks (RNN) is a type of NN structure, which is able to learn the sequential connections and internal states. When RNN has been successfully applied to sequentially learn and predict multiple labels of the data, it typically requires a large number of parameters to observe the above association. Nevertheless RNN with LSTM \cite{hochreiter1997long} is an effective method to exploit label correlation. Researches in different fields also apply RNNs to deal with sequential prediction tasks which utilize the long-term dependency in a sequence, such as image captioning \cite{mao2014deep}, speech recognition \cite{graves2013speech}, language modeling \cite{sundermeyer2012lstm}, and word embedding learning \cite{le2015compositional}. Among multi-label classification, CNN-RNN~\cite{wang2016cnn} is a representative work with promising performance. However, CNN-RNN requires a pre-defined label order for learning, and its limitation to recognize labels corresponding to objects in smaller sizes would be the major concern.

\section{Our Proposed Method}
We first define the goal of the task in this paper. Let $\D=\{(\x_i,\y_i)\}_{i=1}^N = \{\X,\Y\}$ denote the training data, where $\X \in \R^{d\times N}$ indicates a set of $N$ training instances in a $d$ dimensional space. The matrix $\Y \in \R^{\C\times N}$ indicates the associated multi-label matrix, where $\C$ is the number of labels of interest. In other words, each dimension in $\y_c$ is a binary value indicating whether $\x_i$ belongs to the corresponded label~$\c$. For multi-label classification, the goal is to predict the multi-label vector $\hy$ for a test input $\hx$.

\subsection{A Brief Review of CNN-RNN}

CNN-RNN~\cite{wang2016cnn} is a recent deep learning based model for multi-label classification. Since our method can be viewed as an extension, it is necessary to briefly review this model and explain the potential limitations.

As noted earlier, exploiting label dependency would be the key to multi-label classification. Among the first CNN works for tackling this issue, CNN-RNN is composed of a CNN feature mapping layer and a Long Short-Term Memory (LSTM) inference layer. While such an architecture jointly projects the input image and its label vector into a common latent space, the LSTM particularly recovers the correlation between labels. As a result, outputs of multiple labels can be produced at the prediction layer via nearest neighbor search.

Despite its promising performance, CNN-RNN requires a predefined label order for training their models. In addition to the lack of robustness in learning optimal label orders, as confirmed in~\cite{wang2016cnn}, labels of objects in smaller sizes would be difficult to predict if their visual attention information is not properly utilized. Therefore, how to introduce the flexibility in learning optimal label order while jointly exploiting the associated visual information would be the focuses of our proposed work.

\subsection{Order-Free RNN with Visual Attention}
As illustrated in Fig.~\ref{fig:2}, our proposed model for multi-label classification has three major components: feature mapping layer $\fcnn$, attention layer $\fatt$, and LSTM inference layer $\flstm$. The feature mapping layer $\fcnn$ extracts the visual features from the input image $\x_i$ using a pre-trained CNN model. With the attention layer $\fatt$, we would observe a set of feature maps $v_{feat}$, in which each map is learned to describe the corresponding layer of image semantic information. The output of $\fatt$ then goes through the LSTM inference process via $\flstm$, followed by a final prediction layer for producing the label outputs.

During the LSTM inference process, the hidden state vector $\h$ would update the attention layer $\fatt$ with the label inference from the previous time step, guiding the network to visually attend the next region of interest in the input image. Thus, such network designs allow one to exploit \emph{label correlation} using the associated visual information. As a result, the optimal order of label sequences can be automatically observed. In the following subsections, we will detail each component of our proposed model.\\

\subsubsection{Feature Mapping Layer $\fcnn$}

The feature mapping layer $\fcnn$ first extracts visual features $v_{feat}$ by pre-trained CNN models. Following the design in~\cite{liu2016semantic}, we add a fully-connected layer with the output dimension of $c$ after the convolutional layers, which produces the predicted probability $\v_{prob}$ for each label as an additional feature vector. Therefore, the CNN probability outputs can be viewed as preliminary label prediction.

With the ground truth labels given during training (note that positive labels as 1 and negative ones as 0), the learning of $\fcnn$ would update the parameters of the fully-connected layer via observing log-likelihood cross-entropy, while the parameters of the pre-trained CNN remain fixed. By concatenating $m$ feature maps of dimension $k$ in $v_{feat}$, we convert $v_{feat}$ into a single input vector learning visual attention. As a result, the output probability vector of $\fcnn$ can be expressed as follows:
\begin{equation}
\label{eq:feats}
\begin{aligned}
\V = [\V_{feat}, \v_{prob}]
\end{aligned}
\end{equation}

\begin{equation}
\label{eq:V_feat}
\begin{aligned}
\V_{feat} = [\v_1, ..., \v_m], \v_i \in \R^k
\end{aligned}
\end{equation}

\begin{equation}
\label{eq:V_prob}
\begin{aligned} \v_{prob} \in [0, 1]^c.
\end{aligned}
\end{equation}

\subsubsection{Attention Layer $\fatt$}
When predicting multiple labels from an input image, one might suffer from the fact that labels of objects in smaller sizes are not properly identified. For example, \emph{person} typically occupies a significant portion of an input image, while \emph{birds} might appear in smaller sizes and around the corners.

In order to alleviate this problem, we introduce an attention layer $\fatt$ to our proposed architecture, with the goal of focusing on proper image regions when predicting the associated labels. Inspired by Xu et al.~\cite{xu2015show}, who advocated a soft attention-based image caption generator, we advance the same network component in our framework. For multi-label classification, this allows us to focus and describe the image regions of interest during prediction, while implicitly exploiting the label co-occurrence information. In our proposed framework, this attention layer would generate a context vector consisting of weights for each feature map, so that the attended image region can be obtained during each iteration. Later we will also explain that, with such network designs, we can observe optimal label order when learning our RNN-based multi-label classification model.


Following the structure of multi-layer perceptron~\cite{xu2015show}, our attention layer $\fatt$ is conditioned on the previous hidden state $\h_{t-1}$. For each $\v_i$ in Eq.~\ref{eq:V_feat}, the attention layer generates a weight $\alpha_i$, $\alpha_i \in [0, 1]$, which represents the importance weight of feature $i$ in the input image, and predicts the label at this time step. To be more specific, we have:
\begin{equation}
\label{eq:atten_weight}
\begin{aligned}
\epsilon_{i,t} = f_{att}(\v_i, \h_{t-1}) \\
\alpha_{i,t} = \frac{e^{\epsilon_{i,t}} }{ \sum_{j=1}^m e^{\epsilon_{j,t}}},
\end{aligned}
\end{equation}
where $f_{att}$ is the same as the model in~\cite{xu2015show}, and $\h_{t-1}$ would be detailed later in next section.

With $\alpha_{i,t}$, we derive the context vector $\z_t$ with the soft attention mechanism:
\begin{equation}
\label{eq:context}
\begin{aligned}
\z_t = \sum_{i=1}^m \alpha_{i,t}\v_i.
\end{aligned}
\end{equation}

Later, our experiments will visualize and evaluate the contribution of our attention model for multi-label classification.\\

{\color{blue}
\SetKwInOut{Parameter}{Parameter}
\begin{algorithm}[t]
	\label{alg:training}
	\DontPrintSemicolon
	\caption{Training of Our Proposed Model}
	\KwIn{Feature maps $\mathbf{\V}_{feat}$ = [$\mathbf{\v}_1$,...,$\mathbf{\v}_m$] and label vector $\mathbf{\y}$ of an image}
    \Parameter{Resnet fully-connected layer $\mathbf{\theta}_R$, attention layer $\mathbf{\theta}_a$, LSTM layer $\mathbf{\theta}_L$ and prediction layer $\mathbf{\theta}_p$, iteration number $iter$}
	\KwOut{Soft confidence vector $\mathbf{\ty}$}

    \BlankLine
	Randomly initialize parameters\\
    Train $\mathbf{\theta}_R$ by log-likelihood cross-entropy loss and obtain $\mathbf{\v}_{prob}$\\
	\Repeat{$\mathbf{\theta}_a$, $\mathbf{\theta}_L$ and $\mathbf{\theta}_p$ converge}
	{
		\For{$t=1;t \le $iter$;t{+}{+}$}
        {
        	Obtain the context vector $\mathbf{\z}_t$ by \eqref{eq:atten_weight} and \eqref{eq:context}\\
            Obtain the hidden state $\mathbf{\h}_t$ by \eqref{eq:hidden}\\
            Obtain the soft confidence vector $\mathbf{\p}_t$ by \eqref{eq:prob}\\
            Obtain the hard predicted label vector $\mathbf{\tilde{\y}_t}$ by \eqref{eq:max}\\
            Update the candidate pool by \eqref{eq:cand_pool}\\
            Compute the log-likelihood cross-entropy between $\mathbf{\v}_{prob}$ and $\mathbf{\y}$\\
            Perform gradient descent on $\mathbf{\theta}_a$, $\mathbf{\theta}_L$ and $\mathbf{\theta}_p$
            $\mathbf{\v}_{pred}$ = $\mathbf{\p}_t$
        }
    }
\end{algorithm}
} 

\subsubsection{Confidence-Ranked LSTM $\flstm$}
\label{LSTM}
As an extension of recurrent neural network (RNN), LSTM additionally consists of three gate neurons: forget, input, and output gates. The forget gate is to learn proper weights for erasing the memory cell, the input gate is learned to describe the input data, while the output gate aims to control how the memory should be omitted.

In order to exploit and capture the dependency between labels, the LSTM model $\flstm$ in our network architecture needs to identify which label would exhibit a high confidence at each time step. Thus, we concatenate the soft confidence vector $\v_{pred}$ from the previous time step (note that $\v_{pred}$ = $\v_{prob}$ when t=1, and $\v_{pred}$ = $\p_{t-1}$ otherwise), the context vector $\z_t$ and the previous predicted hard label vector $\tilde{\y}_{t-1}$ for deriving the current hidden state vector $\h_t$. This state vector is thus controlled by the aforementioned three gate components. By observing the long-term dependency between labels via the above structure, we can exploit and utilize the resulting label correlation for improved multi-label classification.

We note that, to predict multi-label outputs using LSTM, we pass $\h_t$ through an additional prediction layer consisting of two fully-connected layers, and result in a soft confidence vector $\p_t \in \R^\c$ at time $t$. The hard predicted label $\tl_t = argmax~~\p_t$ indicates the most confident class at the time step $t$, which is then appended to the hard predicted label vector $\tilde{\y}_t$. In the testing phase, by collecting $\tl$ till the ultimate condition, which will be described later in the next section, the final predicted multi-label vector $\tilde{\y}$ can be obtained.

More specifically, we calculate:
\begin{equation}
\label{eq:hidden}
\h_t = f_{LSTM}(\v_{pred}, \z_t, \tilde{\y}_{t-1}, \h_{t-1}),
\end{equation}
where $f_{LSTM}$ denotes the LSTM model. In order to predict $\p_t$, we have:
\begin{equation}
\label{eq:prob}
\p_t = f_{pred}(\v_{pred}, \z_t, \tilde{\y}_{t-1}, \h_t),
\end{equation}

On the other hand, the cross-entropy loss function to minimize at the output layer at time $t$ is:
\begin{equation}
\label{eq:loss}
\L_t = -\sum^{\c}_{i=1} y_i log(\sigma(p_{t,i})) + (1 - y_i)log(1 - \sigma(p_{t,i})),
\end{equation}
where $y_i \in \y$ , $p_{t,i} \in \p_t$, and $\sigma$ is the sigmoid function.

It is worth noting that, the main difficulty of applying LSTM for multi-label classification is its requirement of the ground truth label order during the training process. By simply calculating the cross-entropy between the confidence vector $\p_t$ and the ground truth multi-label vector $\y$, one would not be able to define the order of label prediction for learning LSTM. Moreover, it would be desirable if the label order would reflect semantic dependency between labels presented in training image data.

With the above observation, we view our $\flstm$ in the proposed network architecture as \emph{confidence-ranked LSTM}. Once the previous soft confidence vector $\p_{t-1}$ and hard predicted label vector $\tilde{\y}_{t-1}$ are produced, our model would update $\h_t$ and the attention layer $\fatt$. As a result, we will be able to produce $\p_{t}$ accordingly. In other words, our model achieves the visually attention of objects of semantic interest in the input image, which does not require one to pre-define any specific label order. Therefore, unlike previous works like~\cite{wang2016cnn}, the training of our model does not require the selection of ground truth labels in a predetermined order. Instead, we calculate the loss by comparing the soft confidence vector with the ground truth label vector directly. With our visual attention plus LSTM components, the training process would be consistent with the testing stage. Since the above process relies on visual semantic information for multi-label prediction, one of the major advantages of our model that possible error propagation problems when applying RNN-based approaches can be alleviated.

\subsection{Order-Free Training and Testing}
\subsubsection{Training}

We now explain how our model achieves order-free learning and prediction. As shown in Fig.~\ref{fig:2}, our network design produces outputs of labels ${[\tl_1, ... \tl_T]}$ at time $T$, where each $\tl_i$ denotes the label with the highest confidence at the i-th time step. To avoid duplicate label outputs at different time steps, we apply the concept of candidate label pool as follows.

To initialize the inference process for multi-label learning using our model, the candidate label pool would simply be $\C$ containing all labels. At each time step, the most confident label $\tl_t$ would be selected from the candidate pool, and thus this candidate pool will be updated by removing $\tl_t$ from it. More specifically, for $\tl_t$, we denote it as:
\begin{equation}
\label{eq:max}
\begin{aligned}
\tl_t = arg\max_{\pC_t}~~\p_t, \\
\tilde{\y}_t = \tilde{\y}_{t-1} + \tl_t
\end{aligned}
\end{equation}

\begin{equation}
\label{eq:cand_pool}
\pC_t = \pC_{t-1} - \{\tl_{t-1}\}.
\end{equation}
where $\pC_0$ denotes the full set of labels $\C$, and $\pC_t$ is the set of candidate labels to be predicted at time $t$. From the above label update process, the cardinal of the candidate label set would be subtracted by one at each time step.

\subsubsection{Testing}
We note that, the labels to be predicted during the testing stage can be obtained sequentially using the learned model. However, even with the introduction of the attention layers, prediction error at a time step would be propagated and significantly degrade the prediction performance.

Inspired by~\cite{wang2016cnn}, we apply the technique of \emph{beam search} to alleviate the above problem, and thus the predicting process would be more robust to intermediate prediction errors. More precisely, beam search would keep the best-$K$ prediction paths at each time step. At time step $t+1$, it then searches from all $K\times \c$ successor paths generated from the $K$ previous paths, updates the path probability for all $K\times \c$ successor paths, and maintains the best-$K$ candidates for the following time steps.

In our work, a prediction path represents a sequence of predicted labels with a corresponding path probability, which can be calculated by multiplying the probabilities of all the nodes along the prediction path. At each time step $t$ given a prediction path $[l_1, l_2,\cdots, l_{t-1}]$ and image $I$, its path probability before predicting $l_t$ is calculated as:
\begin{equation}
\begin{aligned}
\label{eq:beam_search}
P_{path} = P(l_1|I) \times P(l_2|I, l_1) \times \cdots \\\times P(l_{t-1}|l_1, l_2, \cdots, l_{t-2}).
\end{aligned}
\end{equation}

Finally, the prediction process via beam search would terminate under the following two conditions:\\
1. The probability output of a particular prediction path is below a threshold (which is determined by cross-validation).\\
2. The length of the prediction path reaches a pre-defined maximum length (which is the largest number of labels in the training set).\\

\section{Experiments}
\subsection{Implementation}
\tabcolsep=1.5pt
\begin{table}[]
	\centering	
		\caption{Evaluation of NUS-WIDE. Note that Macro/Micro P/R/F1 scores are abbreviated as O/C-P/R/F1, respectively. Ours (w/o attention) and Frequency/Rare-first (w/ atten) denote our method with the attention layer removed and using associated pre-defined label orders, respectively.\\}
	
	\begin{tabular}{l||lll|lll}
		\hline
		Method  & C-P  & C-R  & C-F1 & O-P  & O-R  & O-F1 \\ \hline
		KNN		& $32.6$ & $19.3$ & $24.3$ & $43.9$ & $53.4$ & $47.6$ \\
		Softmax & $31.7$ & $31.2$ & $31.4$ & $47.8$ & $59.5$ & $53.0$ \\
		WARP    & $31.7$ & $35.6$ & $33.5$ & $48.6$ & $60.5$ & $53.9$ \\
		CNN-RNN & $40.5$ & $30.4$ & $34.7$ & $49.9$ & $61.7$ & $55.2$ \\
        Resnet-baseline   & $46.5$ & $47.6$ & $47.1$ & $61.6$ & $68.1$ & $64.7$ \\
        Frequency-first (w/ atten)   & $48.9$ & $48.7$ & $48.8$ & $62.1$ & $69.4$ & $65.5$ \\
        Rare-first (w/ atten)   & $53.9$ & $51.8$ & $52.8$ & $55.1$ & $65.2$ & $59.8$ \\
        Ours (w/o atten)   & $60.8$ & $49.5$ & $54.5$ & $68.3$ & $72.4$ & $70.2$ \\
		Ours    & $59.4$ & $50.7$ & $\mathbf{54.7}$ & $69.0$ & $71.4$ & $\mathbf{70.2}$ \\ \hline
	\end{tabular}
	\label{table:nuswide}
\end{table}

To implement our proposed architecture, we apply a ResNet-152~\cite{he2016deep} network trained on Imagenet without fine-tuning, and use the bottom fourth convolution layer for visual feature extraction. We also add a fully-connected layer with dimension of $c$ after the convolutional layer. We employ the Adam optimizer with the learning rate at 0.0003, and the dropout rate at 0.8 for updating $f_{pred}$. We perform validation on the stopping threshold for beam search. As for the parameters for attention and LSTM models, we follow the settings of~\cite{xu2015show} for implementation.

To evaluate the performance of our method and to perform comparisons with state-of-the-art methods, we report results on the benchmark datasets of NUS-WIDE and MS-COCO as discussed in the following subsections.\\ 
\subsection{NUS-WIDE}

NUS-WIDE is a web image dataset which includes 269,648 images with a total of 5,018 tags collected from Flickr. The collected images are further manually labeled into 81 concepts, including objects and scenes. We follow the setting of WARP~\cite{gong2013deep} for experiments by removing images without any label, i.e., 150,000 images are considered for training, and the rest for testing. 

We compare our result with state-of-the-art NN-based models: \textit{WARP}~\cite{gong2013deep} and \textit{CNN-RNN}~\cite{wang2016cnn}. We also also perform several controlled experiments: (1) removing the attention layer, and (2) fixing orders by different methods as suggested by~\cite{jin2016annotation} during training. Frequency-first indicates the labels are sorted by frequency, from high to low, and rare-first is exactly the reverse of frequency-first. The results are listed in Table \ref{table:nuswide}. From this table, we see that our model performed favorably against baseline and state-of-the art multi-label classification algorithms. This demonstrates the effectiveness of our method in learning proper label ordering for sequential label prediction. Finally, our full model achieved the best performance, which further supports the exploitation of visually attended regions for improved multi-label classification.

\begin{figure*}[t!]
	\centering
	\includegraphics[width=0.9\textwidth]{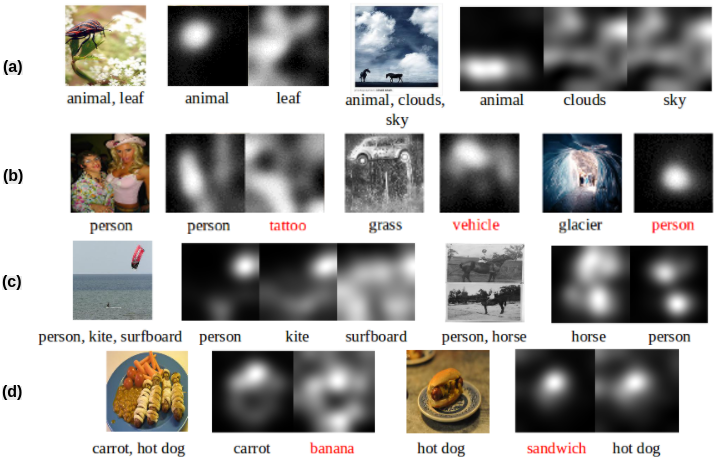}
	\caption{Examples images with correct label prediction in NUS-WISE (a) and MS-COCO (c), those with incorrect prediction are shown in (b) and (d), respectively. For each image (with ground truth labels noted below), the associated attention maps are presented at the right hand side, showing the regions of interest visually attended to. Note that some incorrect predicted labels (in red) were expected and reasonable due to noisy ground truth labels, while the resulting visual attention maps successfully highlight the attended regions.}
	\label{fig:visualization}
\end{figure*}

In Fig. \ref{fig:visualization}(a), we present example images with correct label prediction. We see that our model was able to predict labels depending on what it was actually attended to. For example, since `person' is a frequent label in the dataset, CNN-RNN framework tended to predict it first, because their label order was defined by label occurrence frequency observed during the training stage. In contrast, our model was able to predict animal and horses first, which were actually easier to be predicted based on their visual appearance in the input image. On the other hand, examples of \emph{incorrect} predictions are shown in Fig~\ref{fig:visualization}(b). It is worth pointing out that, as can be seen from these results, the prediction results were actually intuitive and reasonable, and the incorrect prediction was due to the noisy ground truth label. From the above observations, it can be successfully verified that our method is able to identify semantic ordering and visually adapt to objects with different sizes, even given noisy or incorrect label data during the training stage.

\subsection{MS-COCO}

\tabcolsep=1.5pt
\begin{table}[]
	\centering
	
	\caption{Performance comparisons on MS-COCO. Ours (w/o attention) and Ours Frequency/Rare-first (w/ atten) denote our method with the attention layer removed and using associated pre-defined label orders, respectively.\\}
	
	\begin{tabular}{l||lll|lll}
		\hline
		Method  & C-P  & C-R  & C-F1 & O-P  & O-R  & O-F1 \\ \hline
		Softmax & $59.0$ & $57.0$ & $58.0$ & $60.2$ & $62.1$ & $61.1$ \\
		WARP     & $59.3$ & $52.5$ & $55.7$ & $59.8$ & $61.4$ & $60.7$ \\
		CNN-RNN & $66.0$ & $55.6$ & $60.4$ & $69.2$ & $66.4$ & $\mathbf{67.8}$ \\
        Resnet-baseline & $58.3$ & $49.3$ & $53.4$ & $63.9$ & $58.4$ & $61.0$ \\
        Frequency-first (w/ atten)   & $55.8$ & $54.7$ & $55.2$ & $61.4$ & $62.6$ & $62.0$ \\
        Rare-first (w/ atten)   & $59.5$ & $56.5$ & $58.0$ & $57.3$ & $56.7$ & $57.0$ \\
        Ours (w/o atten)   & $69.9$ & $52.6$ & $60.0$ & $73.4$ & $60.3$ & $66.2$ \\
		Ours    & $71.6$ & $54.8$ & $\mathbf{62.1}$ & $74.2$ & $62.2$ & $\mathbf{67.7}$ \\ \hline
	\end{tabular}
	\label{table:mscoco}
\end{table}

MS-COCO is the dataset typically considered for image recognition, segmentation and captioning. The training set consists of 82,783 images with up to 80 annotated object labels. The test set of this experiment utilizes the validation set of MS-COCO (40,504 images), since the ground truth labels of the original test set in MS-COCO are not provided.
,
In the experiments, we compare our model with the \textit{WARP}~\cite{gong2013deep} and \textit{CNN-RNN}~\cite{wang2016cnn} models in Table \ref{table:mscoco}. It can be seen that the full version of our model achieved performance improvements over the Resnet-based baseline by 4.1\% in C-F1 and by 5.6\% in O-F1.

In Figures \ref{fig:visualization}(c) and \ref{fig:visualization}(d), we also present example images with correct and incorrect prediction. It is worth noting that, in the upper left example in Fig. \ref{fig:visualization}(c), although the third attention map corresponded to the label prediction of surfboard, it did not properly focus on the object itself. Instead, it took the surrounding image regions into consideration. Combining the information provided by the hidden state, it still successfully predicted the correct label. This illustrates the ability of our model to utilize both \emph{local} and \emph{global} information in an image during multi-label prediction. 

\section{Conclusion}
We proposed a deep learning model for multi-label classification, which consists of a visual attention model and a confidence-ranked LSTM. Unlike existing RNN-based methods requiring predetermined label orders for training, the joint learning of the above components in our proposed architecture allows us to observe proper label sequences with visually attended regions for performance guarantees. In our experiments, we provided quantitative results to support the effectiveness of our method. In addition, we also verified its robustness in label prediction, even if the training data are noisy and incorrectly annotated.

\bibliographystyle{aaai}
\bibliography{egbib}

\end{document}